\documentclass[10pt,twocolumn,letterpaper]{article}

\usepackage{iccv}
\usepackage{times}
\usepackage{epsfig}
\usepackage{graphicx}
\usepackage{bm}
\usepackage{subcaption}
\usepackage{amsmath}
\usepackage{amssymb}

\usepackage[pagebackref=true,breaklinks=true,letterpaper=true,colorlinks,bookmarks=false]{hyperref}

\iccvfinalcopy 

\def\iccvPaperID{33} 

\ificcvfinal\fi

\begin{document}

\title{Targeted Adversarial Attacks on Generalizable Neural Radiance Fields}

\author{Andr\'as Horv\'ath\\
Pázmány Péter Catholic University\\
Faculty of Information Technology and Bionics\\
{\tt\small horvath.andras@itk.ppke.hu}
\and
Csaba M. J\'ozsa\\
Nokia Bell Labs\\
{\tt\small csaba.jozsa@nokia-bell-labs.com}
}

\maketitle
\ificcvfinal\thispagestyle{empty}\fi

\begin{abstract}
Neural Radiance Fields (NERFs) have recently emerged as a powerful tool for 3D scene representation and rendering. These data-driven models can learn to synthesize high-quality images from sparse 2D observations, enabling realistic and interactive scene reconstructions. However, the growing usage of NERFs in critical applications such as augmented reality, robotics, and virtual environments could be threatened by adversarial attacks.

In this paper we present how generalizable NERFs can be attacked by both low-intensity adversarial attacks and adversarial patches, where the later could be robust enough to be used in real world applications.
We also demonstrate targeted attacks, where a specific, predefined output scene is generated by these attack with success.

\end{abstract}

\section{Introduction}

Neural Radiance Fields (NeRFs) \cite{mildenhall2021nerf}  have emerged as a groundbreaking paradigm in the domain of 3D scene representation and rendering, revolutionizing the way we perceive and interact with virtual environments. NeRFs leverage the power of deep learning to capture intricate scene details \cite{rudnev2022nerf}, enabling the synthesis of photorealistic images from sparse 2D observations \cite{zhang2023transforming}. The ability to reconstruct high-quality scenes from limited input data has propelled NeRFs into the forefront of computer vision, computer graphics, augmented reality \cite{deng2022fov}, robotics \cite{maggio2023loc}, and other related fields.

NeRFs represent 3D scenes as continuous functions, mapping 3D coordinates to their corresponding scene appearance properties, such as color and opacity. This continuous representation distinguishes them from most traditional 3D models, which often rely on discrete voxels or point clouds. In essence, NeRFs can be seen as implicit functions that define the scene's surface, depth and appearance properties, making them particularly suited for complex and detailed scene reconstruction.
They can generate depth maps \cite{deng2022depth} and can be used in navigation \cite{adamkiewicz2022vision}, \cite{xie2023navinerf}, localization \cite{maggio2023loc} and six degrees of freedom orientation estimation \cite{chidananda2022pixtrack}.

The significance of Neural Radiance Fields (NeRFs) lies in their widespread applicability and apart from image rendering, in generating 3D scenes, depth maps, and aiding navigation. However, it is essential to acknowledge that the susceptibility of NeRFs to adversarial attacks can introduce complications and challenges. These attacks could have the potential to produce unrealistic maps and representations, leading to the hallucination of non-existent objects within the scene or the omission of existing objects. As a result, in various applications employing NeRFs, these adversarial perturbations may give rise to erroneous outcomes and hinder accurate scene reconstruction and navigation.

The training process of NeRFs involves capturing multi-view image observations of the scene and optimizing the model to predict accurate color and opacity values for any novel viewing angle within the scene's spatial extent. This approach enables NeRFs to not only render novel viewpoints but also handle dynamic scenes and incorporate additional observations over time. Consequently, NeRFs have opened up exciting possibilities for applications like real-time virtual reality experiences, interactive architectural visualizations \cite{zimny2022points2nerf}, and advanced autonomous robotic systems \cite{adamkiewicz2022vision}.

As NeRFs find increasing adoption in real-world applications, concerns surrounding their vulnerability to adversarial attacks have surfaced. Adversarial attacks aim to exploit vulnerabilities in machine learning models by introducing carefully crafted perturbations to the input data. These perturbations are imperceptible to the human eye but can lead to drastic misclassifications or erroneous predictions.

In their conventional configurations, NeRFs are trained in a scene-specific and object-specific  manner, involving the training of a dedicated neural network for each scene. The neural network's weights store the scene-specific representations and knowledge of views and camera angles. While these networks could potentially be vulnerable to attacks during the training process, exploiting data poisoning \cite{chen2017targeted} or backdoor attacks \cite{kiourti2020trojdrl}, resulting in the production of invalid three-dimensional representations, their lack of generality limits the potential issues in real-world applications. As a consequence, the specialized nature of NeRF training offers a degree of protection against such adversarial perturbations in practical scenarios.

As research on NeRFs has progressed, recent advancements have led to the development of Generalizable Neural Radiance Fields (Gen-NeRFs) \cite{fu2023gen}. These extensions go beyond the original NeRF formulation where scene specific models had to be trained. 
The capabilities of these models can encompass both the generation of novel views and the creation of implicit three-dimensional representations using known previous views and camera poses. Due to the general nature of these methods, there arises a suspicion that they might be susceptible to attacks through perturbations of input pixels in the images. Such attacks could potentially enable the creation of scenes with arbitrary objects.

In this paper, we aim to substantiate this hypothesis by providing a demonstration of the vulnerability of these models to adversarial perturbations on one of the most commonly used Gen-NeRf variant: IBRNet \cite{wang2021ibrnet}, showcasing the potential for generating scenes with arbitrary objects through these attacks.

Adversarial attacks can take various forms within the context of NeRFs, including attacks on the embedded 3D representation, the weights of the trained models or the input pixels. Attacking input pixels is relatively easy, and this method remains the most significant form of attack as it does not require access to the image processing pipeline, making it a potential real-world threat. Consequently, this study focuses on this form of attack by employing targeted attack strategies involving both low-intensity attacks \cite{goodfellow2014explaining} covering all input pixels and patch based attacks \cite{brown2017adversarial} being limited to only a certain region of the image.

Attack strategies can also be distinguished based on  the expected output of the attacks.
In case of untargeted attacks our aim is to modify the output of the network as much as possible, without any restrictions on the output scene of the model.
Meanwhile in case of targeted attack a predefined output scene has to be generated by the model as the result of the attack.

The authors of this paper are aware of only one case where adversarial attacks on NeRFs were investigated. 
In \cite{fu2023nerfool} untargeted attacks have been introduced using Gen-NeRFS. The attack methodology and results are interesting, but untargeted attacks do not pose a substantial real-world threat, as the resulting outputs are often easily detected as non-realistic images, hence their unrestrictedness.

 In contrast, this research delves into targeted attacks, wherein the objective is to create realistic scenes featuring unreal objects, while maintaining the plausibility of the generated depth map or pixel-level output of the original models.
Given the importance of rendered images as the most commonly investigated element, the study specifically focuses on attacking this aspect. By exploring the vulnerability of NeRFs to targeted attacks on the rendered image, our research aims to shed light on potential security risks and the extent of their impact on NERF-based systems. This investigation is expected to provide valuable insights into safeguarding NeRFs against adversarial threats and further enhancing their reliability and practicality in various real-world applications.

In this paper, we embark on a comprehensive exploration of adversarial attacks on NeRFs. We investigate the efficacy of different attack strategies and evaluate their impact on the rendering quality, scene reconstruction accuracy, and generalization capabilities of NeRFs.

Our paper is structured the following way:
in section \ref{SecGENNerf} we briefly describe Generalized Neural Radiance Fields,
 in section \ref{SecAdvAttack} we introduce the most commonly applied adversarial attack methodologies and algorithms, in section \ref{SecResults} we describe our experiments and results and in section \ref{SecConclusion} we draw conclusion from them.

\section{Generalizable NeRfs}\label{SecGENNerf}

NeRFs present a cutting-edge approach in leveraging deep neural networks to generate 3D representations of objects or scenes from 2D images. This innovative technique involves encoding the complete object or scene within an artificial neural network, which then predicts the light intensity, also known as radiance, at any specific point in the 2D image. As a result, NeRFs enable the creation of novel 3D views from various angles, revolutionizing the generation of highly realistic 3D objects automatically.

The exceptional potential of NeRFs lies in their capacity to represent 3D data more efficiently compared to other existing methods. This efficiency opens new avenues for generating highly realistic 3D objects with remarkable promise. Moreover, when combined with complementary techniques, NeRFs offer the exciting prospect of significantly compressing 3D representations of the world, reducing data sizes from gigabytes to mere tens of megabytes \cite{deng2023compressing}. Such advancements hold significant implications for various fields, enabling streamlined and versatile 3D data generation and manipulation.

Gen-NeRF variants like \cite{chen2021mvsnerf},\cite{liu2021neural}, \cite{reizenstein2021common}, 
\cite{wang2021ibrnet} enable cross-scene generalization via two modifications on top of traditional NeRFs:
 Firstly, these variants condition NeRFs on the source views of new scenes. This involves utilizing a limited number of observed source views from a new scene to extract features via a Convolutional Neural Network (CNN) encoder. These features are then used as scene priors and fed into a vanilla NeRF model as inputs. Secondly, the variants incorporate a ray transformer, which operates on all points along the same ray, enhancing the density prediction.

From the various variants of Gen-NeRFs we have selected IBRNet \cite{wang2021ibrnet} for our investigations, which is commonly applied and highly cited variant, capable of rendering state of the art images from new views on novel scenes.
 
To render a pixel in the Image Based Network (IBRNet), the following steps are involved:
\begin{itemize}
    \item Step 0: 2D feature maps ${W_i}^S_{i=1}$ are calculated from a total of $S$ source views ${I_i}^S_{i=1}$ using a CNN encoder $E$, where $W_i= E(I_i)$ represents a 3D tensor. Notably, this process requires only a one-time effort for each new scene.
    \item Step 1: A ray $r(t) = o + td$ is emitted from the origin $o$ along the view direction $d$ to pass through the pixel to be rendered. 3D points ${x_k}$ are sampled along the ray based on an ordered depth sequence ${t_k}$ drawn from a certain distribution.
    \item Step 2: Each sampled 3D point ${x_k}$ is projected onto the image planes of source views using a project transformation $\pi$, obtaining the corresponding scene features $[W_i] \pi(x_k)$ for all $S$ source views.
    \item Step 3: The scene features acquired in the previous step are applied to an MLP model $f$ to derive the color $c_k$ and density feature $f_k^\sigma$ for each point.
    \item Step 4: The density features of all sampled points along the ray are fed into a ray transformer $T$, yielding the predicted density $\sigma_k$ for each point.
    \item  Step 5: Volume rendering is performed combining the predicted colors and densities for every rendered pixel.
\end{itemize}

During training, the networks E, f, and T are updated using the Mean Squared Error (MSE) loss or other pixel-based distance metrics, ensuring effective learning of the rendering process. 

\begin{figure*}[htp]
\centering
\includegraphics[width=6.0in]{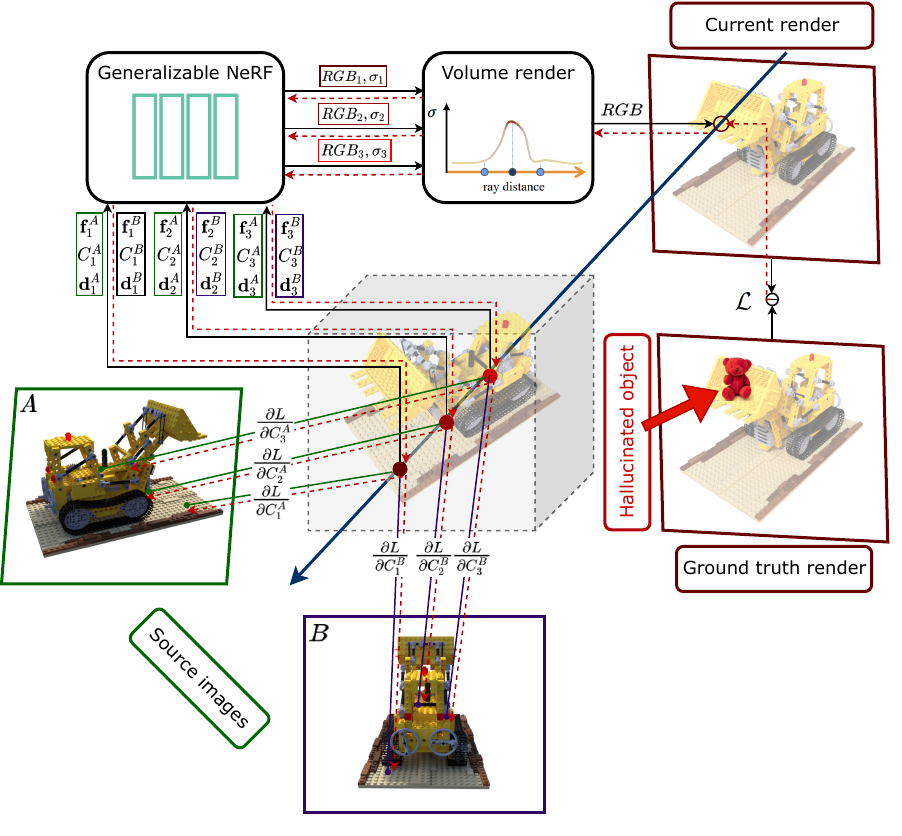}
\caption{
This figure depicts the architecture of IBRNet. The architecture contains and encoder ($E$), multilayered networks for color generation ($f$) and a ray transformer network ($T$). These elements are all fully differentiable, this way gradients can be back-propagated from a final ground truth image and even the input pixels can be altered in a desired manner. The source of the original image is \cite{wang2021ibrnet}.}
\label{fig:ibrnet}
\end{figure*}

The whole architecture of IBRNet is depicted in Figure \ref{fig:ibrnet}. This whole process is differentiable, therefore once a model is trained, the error can be derived back to the pixels, and freezing the model weights the input pixels, or parts of the input pixels can be trained.
We have used a pretrained model, which was trained on multiple datasets simultaneously (LLF \cite{mildenhall2019llff}, RealEstate 10k \cite{zhou2018stereo}, Google Scanned Objects \cite{downs2022google}, etc. ) to be able to cope with generic scenes.
For the sake of reproducibility, the same pretrained model and data for training and evaluation are available at:  \url{https://drive.google.com/drive/folders/1qfcPffMy8-rmZjbapLAtdrKwg3AV-NJe}.

There are more recent implementations  and variants of Gen-Nerfs, such as \cite{johari2022geonerf}, which apply geometric constraints to be more efficient, or \cite{fu2023gen} where even hardware constraints were considered, but these approaches do not differ significantly from the model of our selection, therefor we believe that the attacks presented here can be generalized for these variants as well.

Gen-Nerfs represent a highly promising real-world solution for novel view synthesis, owing to their remarkable ability to generalize across different scenes, facilitating instant rendering on previously unseen environments. Despite the critical significance of adversarial robustness in practical applications, limited attention has been given to exploring its implications specifically for Gen-Nerf. We postulate that Gen-Nerf's conditioning on source views from new scenes, often sourced from the Internet or third-party providers, may introduce novel security concerns in real-world scenarios. Additionally, the conventional understanding and solutions for achieving adversarial robustness in neural networks may not directly apply to Gen-Nerfs, given its distinctive 3D nature and diverse operations.

\section{Adversarial attacks}\label{SecAdvAttack}

The concept of adversarial attacks originated from the pioneering work of \cite{szegedy2013intriguing}. It brought to light a crucial revelation about deep neural networks. Despite their ability to generalize effectively and perform well on conventional input data and even on similar inputs, they possess a vulnerability to exploitation by malicious agents. This vulnerability stems from the high-dimensional nature of inputs, enabling the generation of non-realistic input samples that generate outputs, which deviate drastically from human judgment and the expected outcomes.

The initial adversarial attacks proposed by Goodfellow et al. \cite{goodfellow2014explaining} involved calculating the sign of the gradient of the cost function ($J$) with respect to the input ($\bm{x}$) and expected output ($y$), which was then scaled by a constant ($\epsilon$) to control the intensity of the noise. This method, known as the Fast Gradient Sign Method (FGSM), allowed for rapid generation of attacks.

Rozsa et al. \cite{rozsa2016adversarial} extended FGSM by utilizing not just the sign of the raw gradient but also a scaled version of the gradient's magnitude, termed the Fast Gradient Value method.

Furthermore, Dong et al. \cite{dong2018boosting} proposed an iterative version of FGSM that incorporated momentum into the equation. The inclusion of momentum was inspired by the concept of optimization during model training, with the goal of avoiding poor local minima and non-convex patterns in the objective function's landscape.

Moosavi et al. \cite{moosavi2016deepfool} approached adversarial attacks from the perspective of binary classifier robustness. They formulated the idea that a binary classifier's robustness at a given point $\bm{x}_0$ is determined by its distance from the separating hyperplane $\Delta(\bm{x}_0;f)$. They derived a closed-form formula to calculate the smallest perturbation required to change the classifier's output and applied these perturbations iteratively to the image until the classifier's decision changed. This approach was later extended to address multiclass classification problems as well.

While these methods were crucial for theoretical understanding, their application to neural networks in practical, real-world applications has limited significance due to their low-intensity, constrained noise application. In real-world scenarios, even the smallest perturbations, such as those arising from environmental factors like perspective, illumination changes, or lens distortion, can completely disrupt the desired results. Therefore, the utilization of these attacks in practical applications is not feasible \cite{lu2017no}.

In \cite{brown2017adversarial}, \cite{athalye2017synthesizing} robust and real-world attacks were presented against various classification networks. These methods create an adversarial patch, where instead of the global, but low-intensity approaches, distortions appear in a region with limited area, but intensity values are not bounded\footnote{apart from the global bounds of image values}. 
Successful attacks with adversarial patches were also demonstrated using black and white patches only \cite{eykholt2017robust}, where not the intensities of the patch, but the locations and sizes of the stickers are optimized. These attacks, where the gradients of the networks are not necessarily used during optimization open space towards black-box attacks \cite{alzantot2018genattack}, \cite{papernot2017practical}, where the attacker needs access only to the final responses, confidence values to generate attacks using evolutionary algorithms.

A general overview of adversarial attacks, containing a more detailed description of most of the previously mentioned methods can be found in the following survey paper \cite{akhtar2018threat}.
The resilience of segmentation networks against adversarial attacks was investigated heavily in the past years \cite{xie2017adversarial}, \cite{metzen2017universal}, \cite{arnab2018robustness}, \cite{al2021class}.

Subsequent years witnessed extensive investigations into the potential of exploiting adversarial attacks. Researchers developed novel attack strategies to enhance the robustness of generated attacks \cite{brown2017adversarial}, \cite{athalye2017synthesizing}, even enabling black-box attacks, which do not require access to the network gradients \cite{eykholt2017robust}, \cite{alzantot2018genattack}, \cite{papernot2017practical}.

Moreover, advancements were made in extending adversarial attacks to more complex tasks beyond classification, such as detection and localization problems \cite{thys2019fooling}. These innovative techniques were applied to diverse network architectures, including Faster-RCNN \cite{chen2018shapeshifter}.

According to our best knowledge adversarial attacks has not be presented and investigated in Gen-Nerf models apart from \cite{fu2023nerfool}, which is restricted to low-intensity and untargeted attacks.

\section{Experiments and Results}\label{SecResults}

We have selected a pretrained model of IBRNet and manually created modified images which were serving as ground truths for the attacks. These images were obtained first as the actual outputs of the networks and the modifications were done using GIMP.

Since these modifications were manual we have to admit the they can be biased in two ways. On one hand they might disturb the real structure of the images (artificial insertion and deletion might cause extremely strong edges in the image), on the other hand the modifications are subjective and other people might desire different modifications. We would argue that this subjectivity is unavoidable and we were carefully generating three different types of modifications:
\begin{itemize}
    \item types where the shape of existing objects are modified.
    \item types where existing images were deleted from scenes and substituted by background pixels
    \item types where new objects were added to the scenes
\end{itemize}

A few samples of these modifications and the result of attacks using these image as desired outputs can be seen in Figure \ref{fig:attacksamples}.

\subsection{Low-intensity Attacks}

\begin{figure*}[htbp]
    \centering
    \begin{subfigure}[b]{.32\textwidth}
        \centering
        \includegraphics[width=\linewidth]{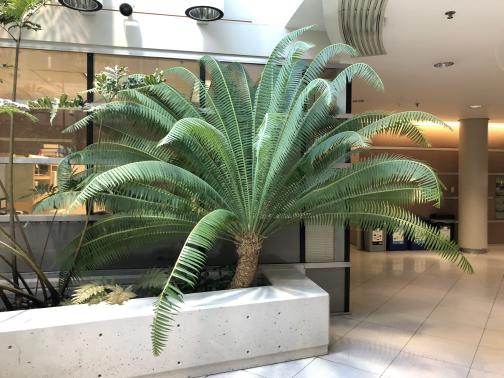}
    \end{subfigure}%
    \hfill
    \begin{subfigure}[b]{.32\textwidth}
        \centering
        \includegraphics[width=\linewidth]{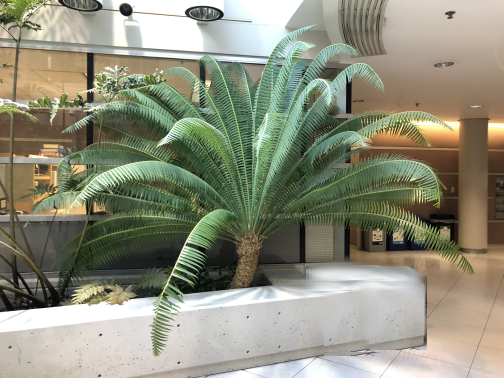}
    \end{subfigure}%
    \hfill
    \begin{subfigure}[b]{.32\textwidth}
        \centering
        \includegraphics[width=\linewidth]{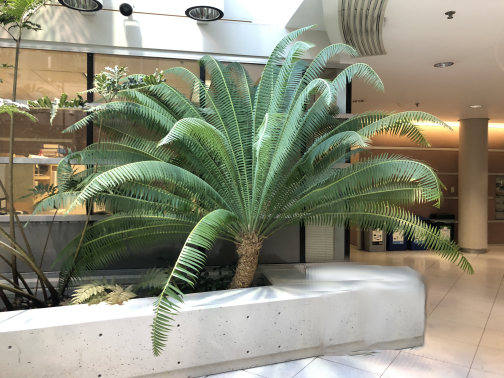}
    \end{subfigure}\\
    \begin{subfigure}[b]{.32\textwidth}
        \centering
        \includegraphics[width=\linewidth]{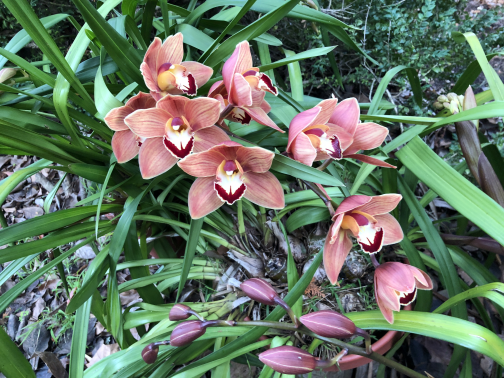}
    \end{subfigure}%
    \hfill
    \begin{subfigure}[b]{.32\textwidth}
        \centering
        \includegraphics[width=\linewidth]{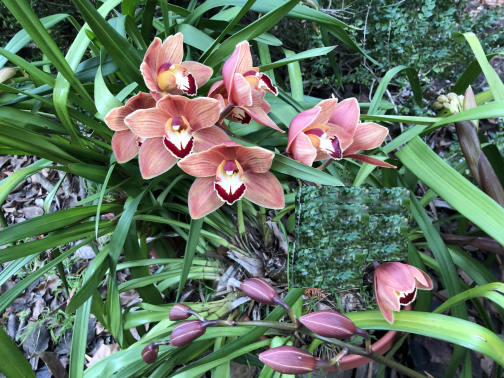}
    \end{subfigure}%
    \hfill
    \begin{subfigure}[b]{.32\textwidth}
        \centering
        \includegraphics[width=\linewidth]{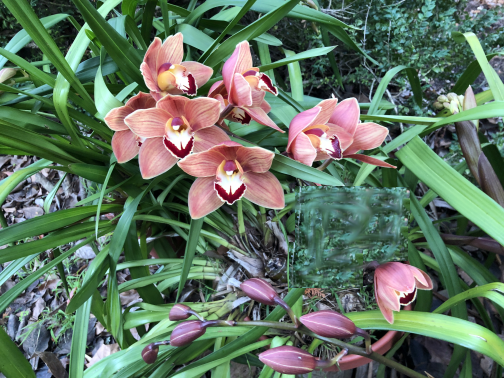}
         \end{subfigure}
        \\
    \begin{subfigure}[b]{.32\textwidth}
        \centering
        \includegraphics[width=\linewidth]{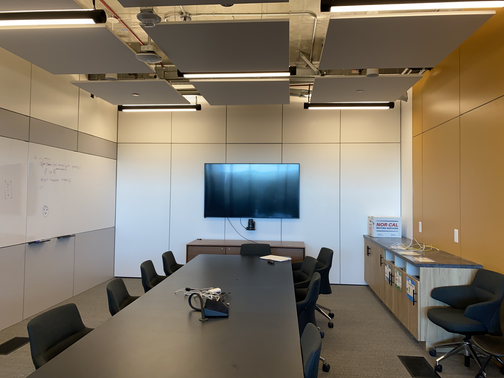}
    \end{subfigure}%
    \hfill
    \begin{subfigure}[b]{.32\textwidth}
        \centering
        \includegraphics[width=\linewidth]{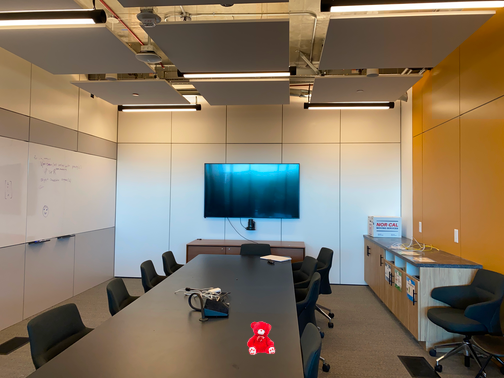}
    \end{subfigure}%
    \hfill
    \begin{subfigure}[b]{.32\textwidth}
        \centering
        \includegraphics[width=\linewidth]{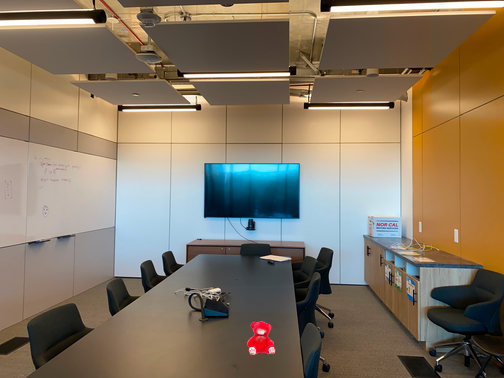}
    \end{subfigure}
    \caption{Samples cases from evaluation part of the LLFF dataset. Here we display three different samples in each row. The first column contains the original output images of the network without any attacks, the second column contains the modified image which were used as ground truth during the attacks. These modifications were done manually. The third column contains the output image of the network after the attack. In this setup the images were generated using ten different view and attacks were applied on all input images. The attacks were generated using FGSM for 1000 iterations with and $\epsilon$ value of $0.01$.
    As these images demonstrate adversarial attacks were successful and we were able to modify objects in the scene (fern), delete objects from the scene (orchids) and render non-existing objects in the scene (room).}
    \label{fig:attacksamples}
\end{figure*}

For our investigation into low-intensity attacks, we opted for the iterative version of FGSM with momentum \cite{dong2018boosting} as the attack mechanism. Our setup involved 1000 iterations, with parameter $\epsilon$ set to $0.01$.

In a typical low-intensity attack on classification problems, a single input image is used, allowing modifications to all its pixels until a predefined threshold is reached. However, since Gen-NeRFs utilize multiple input images, referred to as source images or source views, attackers can simultaneously modify all or a subset of these images. To explore the impact of different attack scenarios, we devised five setups with varying numbers of source images: 10, 8, 6, 5, and 4. The quality of the generated image depends on the number of source images, generally improving with an increase in this number. Our investigation covered cases where one, two, three, and so on, up to all source images were subject to modification.

This investigation holds significance as it addresses real-world scenarios where images from events are uploaded to a common dataset by users or multiple autonomous robots. In such cases, understanding the necessary number and percentage of images to be attacked for successful modifications in the rendered output image becomes crucial. This way Our research aims could provide valuable insights into enhancing the security and reliability of Gen-NeRFs in various practical applications.

\begin{figure}[htbp]
    \centering
    \begin{subfigure}[b]{0.5\textwidth}
        \centering
        \includegraphics[width=\linewidth]{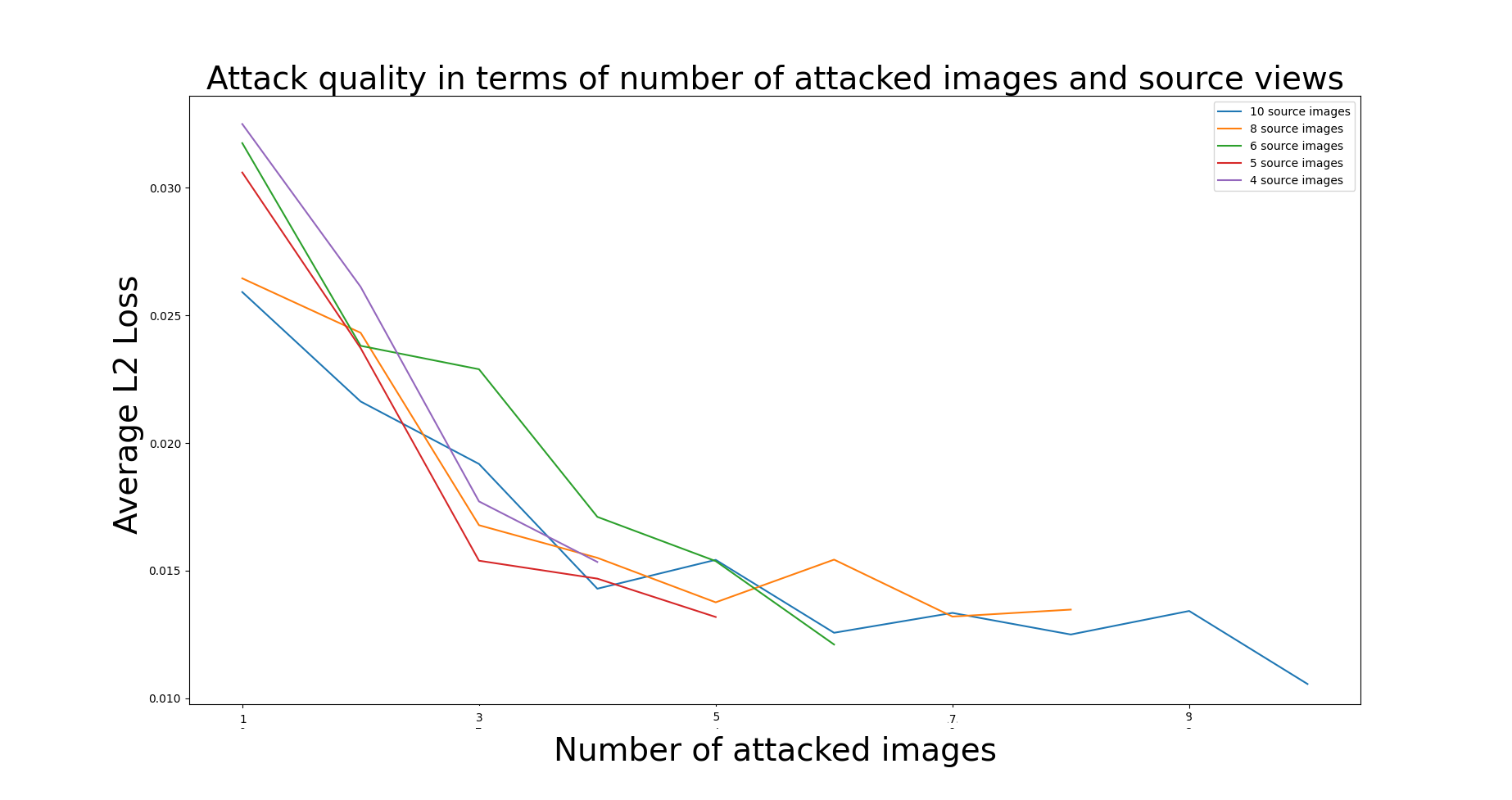}
    \end{subfigure}
    \caption{This plot depicts the dependence of attack quality on the number of source views and the number of attacked samples in case of Gen-Nerfs. The Y axes plots the average $\ell_2$ distance between the pixels of the ground truth image and the image generated by the network after the attack. Lower values mean the attack was more successful, since this case the network output was closer to our desired output. The X axes contains the number of attacked images, meanwhile the different colored plots depicts outputs generated from different number of source views. As it can be seen from these results attacks are not successful (they generate a larger distance) until the number of attacked views will not reach the majority of the source views. Each point in these measurements were generated as the average of 10 independent runs and on ten different scenes.}
    \label{fig:attackqulaity}
\end{figure}

The quality of the attack was measured as the average $\ell_2$ distance between the generated image and our hand-modified ground truth image. 
We have executed this experiment on ten different scenes, repeating each attack ten times (to average out the stochastic nature of the attack algorithm) and the quantitative summary of the results can be seen in Figure \ref{fig:attackqulaity}.

These results clearly indicate that attacks were successful in most cases when a significant majority of the source views were targeted. In our setup, an attack can be considered successful when the average pixel distance dropped below $0.015$, while unsuccessful attacks resulted in values above $~0.020$. It is important to note that these threshold values may vary depending on the scene, but as observed in Figure \ref{fig:attacksamples}, scenes with only a small region altered in the image can be used as a rule of thumb.

These findings highlight the overall robustness of Gen-Nerfs, as the generated images remained reliable in cases where the majority of the source images were left untouched. However, the study also underscores the vulnerability of the system when an attacker gains access to most of the source images, enabling arbitrary modifications to the output. Understanding and addressing these security implications are crucial as Gen-NeRFs and similar technologies advance, ensuring their safe and reliable application in various practical scenarios.

\subsection{Patch-based attacked}

Low-intensity attacks may hold academic interest, but their significance diminishes when considering real-world applications, primarily due to the limited access attackers have to the image processing pipeline. However, the most straightforward and practical way to target neural networks is by modifying the real environment itself. In such scenarios, attackers can manipulate small regions within the image while freely altering the pixel values in this designated region. To effectively simulate and study these real-world threats, we have focused our investigation on patch-based attacks.

Patch-based attacks provide a suitable framework to understand the vulnerabilities of neural networks in the face of real-world adversarial manipulations. By restricting our attention to specific regions in the image, we emulate the scenario where an attacker can locally modify the environment while leaving the rest of the scene intact. The arbitrary nature of pixel values within these patches allows us to evaluate the robustness of the neural networks against unpredictable and potentially damaging alterations.

For low-intensity attacks, the algorithm's crucial parameter is the $\epsilon$ value, intended to ensure the challenging detectability of these modifications.
Similarly the size of the patch applied is the most crucial parameter in patch-based attacks, akin to the significance of the amount of maximal change in low-intensity attacks. To examine the impact of patch size on these attacks, we employed the same set of 10 scenes previously generated. For each scene, desired attack outputs were manually specified, and patches were automatically placed at the center of the images. This approach ensured that the patches were not closely positioned to the regions already modified.

Clearly, a patch covering the modified region could influence the outcome, especially when applied near or at the boundary of the effect. However, the most critical scenario to consider is when patches have far-reaching effects, altering pixels that are not in close proximity to them and keeping the original output value of other regions.

\begin{figure}[htbp]
    \centering
    \begin{subfigure}[b]{0.5\textwidth}
        \centering
        \includegraphics[width=\linewidth]{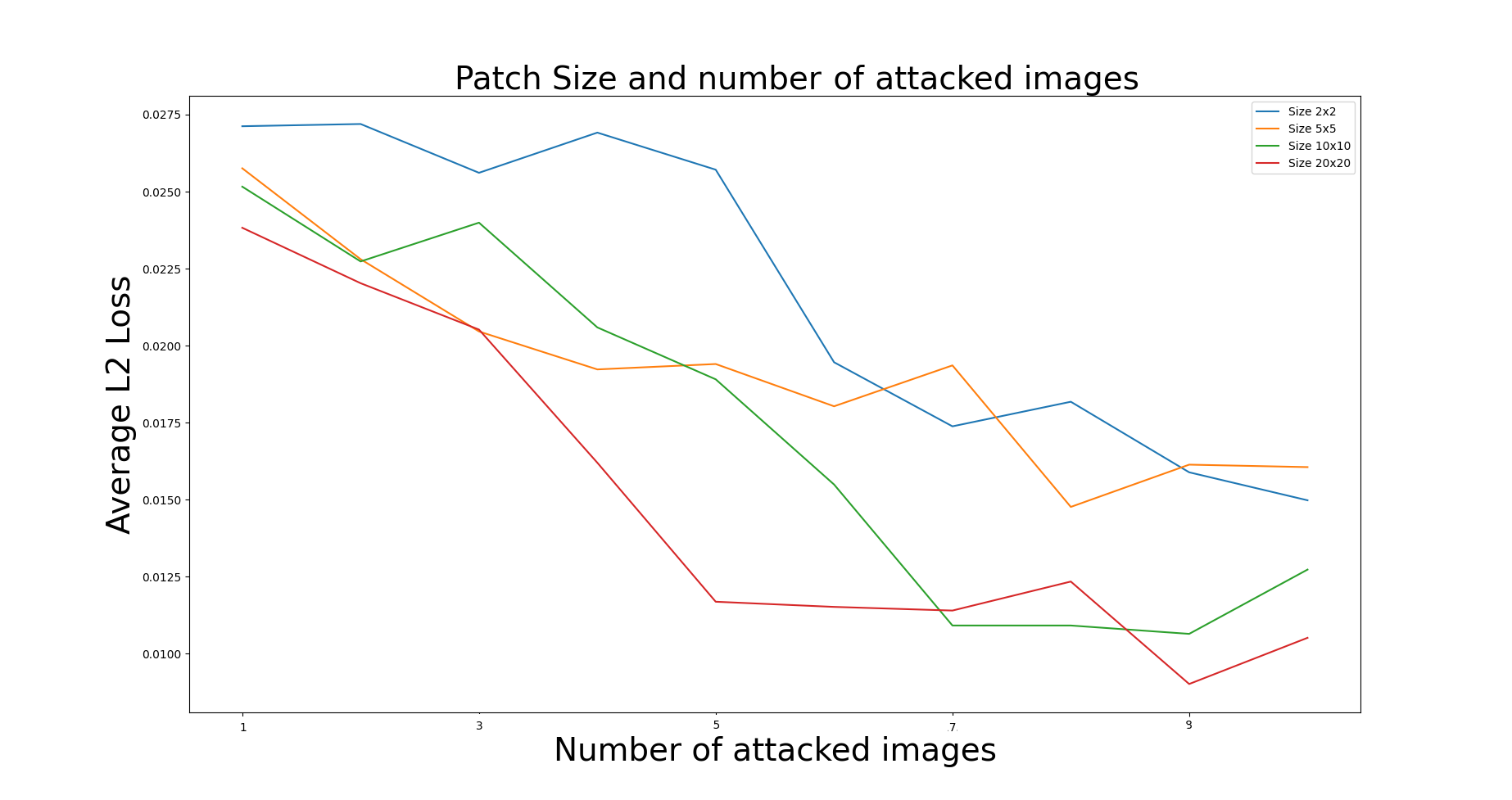}
    \end{subfigure}
    \caption{This plot illustrates how the attack quality is influenced by the number of source views and the size of the patch applied during the attack. The Y-axis represents the average $\ell_2$ distance between the pixels of the ground truth image and the image generated by the network after the attack. Lower values indicate more successful attacks, as they result in the network output being closer to our desired output. On the other hand, the X-axis represents the number of attacked images, while the various colored plots depict outputs generated from different patch sizes.
The results demonstrate that attacks are not successful (they generate a larger distance) until the number of attacked views encompasses the majority of the source views or when the patch size is too small. These measurements are based on the average of 10 independent runs and conducted on ten different scenes}
    \label{fig:patchsize}
\end{figure}

Our experimental investigations involved generating patches of sizes $2\times2$, $5\times5$, $10\times10$, and $20\times20$, and then assessing their respective effects on the scenes. The results of these experiments are illustrated in Figure \ref{fig:patchsize}, providing valuable insights into the relationship between patch size and the success of patch-based attacks. 

The results clearly demonstrate the feasibility of patch-based attacks when the patch size is sufficiently large (typically $10\times10$ patches in our experiments) and when the patches are prevalent in the majority of images. In our investigations, utilizing ten source views, attacks were generally successful if at least four of them contained a patch large enough to cause significant impact.

It is essential to highlight that in this experiment, the patches were independently optimized for each source image. Consequently, the pixel values at the same location could differ across different images, enabling the attacker to tailor their patches specifically to exploit the vulnerabilities in each individual source view.

These findings underscore the potential threat posed by patch-based attacks and emphasize the importance of developing robust defenses against such manipulation techniques. Understanding the adaptability of these attacks to various scenarios is crucial for strengthening the security of neural network systems in real-world applications.

These preliminary results demonstrate the feasibility of patch-based attacks on Gen-NeRFs. However, our simulations deviate from real-world setups in the following aspects:

\begin{itemize}
    \item The patches are consistently positioned at the center of images, and their locations remain unchanged regardless of the viewpoint.
    \item The pixel values within the patches are optimized independently from each other and can vary across different input images.
\end{itemize}
c
Addressing these differences in the future is essential to simulate scenarios where an attacker introduces a real object into a scene. Despite these limitations, we are optimistic that this research paves the way for real-world adversarial applications, such as stickers on Gen-NeRFs.

In this section, we include solely the essential parameters that constitute our algorithm. However, to ensure reproducibility, all the experiment scripts in PyTorch will be openly shared on GitHub in the final version. Additionally, these scripts can be accessed in the supplementary material of the current version, maintaining the anonymous nature of the double-blind review process. The CMT portal for the workshop accepts only doc/pdf format so the uplaoded file was renamed to a pdf, but it is indeed a zip.

\section{Conclusion}\label{SecConclusion}
We have demonstrated  targeted adversarial attacks on Gen-Nerf, revealing important insights into the security vulnerabilities of these networks. The success of the attacks, utilizing methods commonly employed in classification tasks, emphasizes the ease with which malevolent attackers can manipulate the generated images. However, our findings also demonstrate the relative robustness of Nerfs when multiple views are utilized and not all source images are accessible to the attacker. In such cases, the effectiveness of the attack diminishes, indicating the importance of safeguarding access to critical source images. In cases where the attacker has access to the majority of the view the quality of the attacks increases significantly.

Additionally, we explored patch-based attacks, where limited regions of the image are targeted, but arbitrary values can be introduced. Remarkably, these attacks are not restricted to local neighborhoods, as even distant regions can be manipulated with such modifications. The position and view angle of these patches proved to have little impact on their efficacy, further accentuating the potential threat posed by these attacks.

While our results indicate that these attacks have the potential to be robust enough for real-world applications, it is essential to acknowledge that further investigations are necessary to fully comprehend their implications and develop effective countermeasures. As the field of NERFs continues to advance, addressing security concerns and improving defenses against adversarial attacks becomes imperative to ensure the trustworthy deployment of these technologies in various domains.

\section*{Acknowledgement}\label{ACK}
This research has been partially supported by the Hungarian Government by the following grants: 2018-1.2.1-NKP00008: Exploring the Mathematical Foundations of Artificial Intelligence and TKP2021\_02-NVA-27 – Thematic Excellence Program. The support of the Alfréd Rényi Institute of Mathematics if also gratefully acknowledged.

{\small
\bibliographystyle{ieeetr}
\bibliography{egbib}
}

\end{document}